\documentclass[letterpaper]{article} 
\usepackage{aaai23}  
\usepackage{times}  
\usepackage{helvet}  
\usepackage{courier}  
\usepackage[hyphens]{url}  
\usepackage{graphicx} 
\usepackage{natbib}  
\usepackage{caption} 
\usepackage{multirow}
\usepackage{amsmath}
\usepackage{amsfonts}
\usepackage{booktabs}
\usepackage{threeparttable}
\usepackage{nicefrac}
\usepackage[symbol]{footmisc}
\usepackage[justification=centering]{caption}
\usepackage{algorithm}
\usepackage{algorithmic}

\usepackage{newfloat}
\usepackage{listings}
\DeclareCaptionStyle{ruled}{labelfont=normalfont,labelsep=colon,strut=off} 
\floatstyle{ruled}
\newfloat{listing}{tb}{lst}{}
\floatname{listing}{Listing}
\title{Spatio-Temporal Meta-Graph Learning for Traffic Forecasting}
\author{
    Renhe Jiang\textsuperscript{*\rm 1,2},
    Zhaonan Wang\textsuperscript{*\rm 2},
    Jiawei Yong\textsuperscript{\rm 3},
    Puneet Jeph\textsuperscript{\rm 2},
    Quanjun Chen\textsuperscript{\rm 2},   \\
    Yasumasa Kobayashi\textsuperscript{\rm 3},
    Xuan Song\textsuperscript{\rm 2},
    Shintaro Fukushima\textsuperscript{\rm 3},
    Toyotaro Suzumura\textsuperscript{\rm 1}
}
\affiliations{
    \textsuperscript{\rm 1} Information Technology Center, The University of Tokyo \\
    \textsuperscript{\rm 2} Center for Spatial Information Science, The University of Tokyo \\ 
    \textsuperscript{\rm 3} Toyota Motor Corporation \\
    \{jiangrh, znwang, songxuan\}@csis.u-tokyo.ac.jp;
     \{jiawei\_yong, s\_fukushima\}@mail.toyota.co.jp

}

\usepackage{bibentry}
\begin{document}
\maketitle
\footnotetext{* Corresponding authors who contributed equally.} 

\begin{abstract}
Traffic forecasting as a canonical task of multivariate time series forecasting has been a significant research topic in AI community. To address the spatio-temporal heterogeneity and non-stationarity implied in the traffic stream, in this study, we propose Spatio-Temporal Meta-Graph Learning as a novel Graph Structure Learning mechanism on spatio-temporal data. Specifically, we implement this idea into Meta-Graph Convolutional Recurrent Network (MegaCRN) by plugging the Meta-Graph Learner powered by a Meta-Node Bank into GCRN encoder-decoder. We conduct a comprehensive evaluation on two benchmark datasets (i.e., METR-LA and PEMS-BAY) and a new large-scale traffic speed dataset called EXPY-TKY that covers 1843 expressway road links in Tokyo. Our model outperformed the state-of-the-arts on all three datasets. Besides, through a series of qualitative evaluations, we demonstrate that our model can explicitly disentangle the road links and time slots with different patterns and be robustly adaptive to any anomalous traffic situations. Codes and datasets are available at \url{https://github.com/deepkashiwa20/MegaCRN}.
\end{abstract}

\section{Introduction}

Spatio-temporal data, streamed by sensor networks, are widely studied in both academia and industry given various real-world applications. Traffic forecasting \cite{yu2018spatio, li2018diffusion, zheng2020gman, bai2020adaptive, lee2021learning}, as one canonical task, has been receiving increasing attention with rapid developing Graph Convolutional Networks (GCNs) \cite{defferrard2016convolutional, kipf2016semi, velivckovic2017graph}. This spatio-temporal modeling task can be formulated similarly to multivariate time series (MTS) forecasting \cite{wu2020connecting, cao2020spectral, shang2021discrete}, but with extra prior knowledge from the geographic space (\textit{e.g.} sensor locations, road networks) to imply the dependency among sensor signals. Compared with ordinary MTS, traffic data (\textit{e.g.} traffic speed and flow) potentially contain \textbf{spatio-temporal heterogeneity}, as traffic condition differs over roads (\textit{e.g.} local road, highway, interchange) and time (\textit{e.g.} off-peak and rush hours). Moreover, \textbf{non-stationarity} makes the task even more challenging when X factors, including accident and congestion, present.

\begin{figure}[t]
	\centering
	\includegraphics[width=1\linewidth]{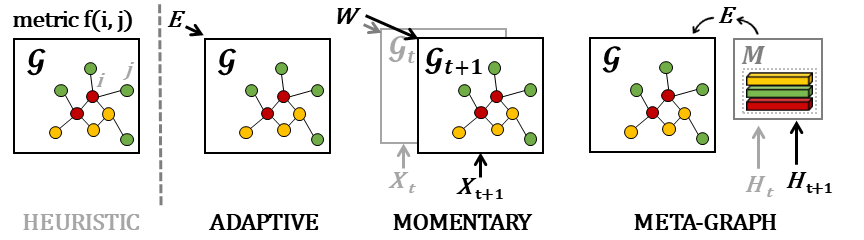}
	\caption{Progression of Graph Structure Learning for Spatio-Temporal Modeling}
	\vspace{-12pt}
	\label{fig:intro}
\end{figure}

For more effective traffic forecasting, the existing works have made tremendous progress by modeling latent spatial correlation among sensors and temporal autocorrelation within time series. Since these two relationships can be naturally represented by graph and sequence respectively, the mainstream models handle them by leveraging GCN-based modules \cite{diao2019dynamic, guo2019attention, geng2019spatiotemporal, zhang2021traffic} and sequence models, such as Recurrent Neural Networks (RNNs) \cite{li2018diffusion, bai2020adaptive, ye2021coupled}, WaveNet \cite{wu2019graph}, Transformer \cite{zheng2020gman, wang2020traffic, xu2020spatial}. Particularly, to perform convolution-like operations on graphs, GCNs require an auxiliary input that characterizes the topology of the underlying spatial dependency. This essential part is defined based on certain metrics in early works, such as inverse-distance Gaussian kernel \cite{yu2018spatio, li2018diffusion}, cosine similarity \cite{geng2019spatiotemporal}. 

However, this pre-defined graph not only relies on empirical laws (\textit{e.g.} Tobler's first law of geography) which does not necessarily indicate an optimal solution, but ignores the dynamic nature of traffic networks. This twofold limitation has stimulated explorations in two lines of research. The first one aims to find the optimal graph structure that facilitates the forecasting task. GW-Net \cite{wu2019graph} pioneers along this direction by treating the adjacency matrix as free variables (\textit{i.e.} parameterized node embedding $E$) to train, which generates a so-called \textit{adaptive} graph (in Figure \ref{fig:intro}). Models including MTGNN \cite{wu2020connecting}, AGCRN \cite{bai2020adaptive}, GTS \cite{shang2021discrete} fall into this category, integrating MTS and traffic forecasting with Graph Structure Learning (GSL) \cite{zhu2021deep}. In the other line of research, attempts have been made to tackle network dynamics using matrix or tensor decomposition \cite{diao2019dynamic, ye2021coupled} and attention mechanisms \cite{guo2019attention, zheng2020gman}. Motivated by GSL, recent models like SLCNN \cite{zhang2020spatio}, StemGNN \cite{cao2020spectral} further try to learn a time-variant graph structure from observational data. The spatio-temporal graph (STG) derived in this way is essentially input-conditioned, in which parameters denoted by $W$ project observations into node embeddings (termed as \textit{momentary} graph in Figure \ref{fig:intro}).

Thus far, while spatio-temporal regularities have been studied systematically, spatio-temporal heterogeneity and non-stationarity have not been tackled properly. Although the heterogeneity issue can be alleviated to some extent by applying attentions over the space and time \cite{guo2019attention, zheng2020gman}, sensor signals of different natures are still left entangled, not to mention that incidents are simply untreated. Therefore, we are motivated to propose a novel spatio-temporal meta-graph learning framework. The term \textit{meta-graph} is coined to describe the generation of node embeddings (similar in \textit{adaptive} and \textit{momentary}) for GSL. Specifically, our STG learning consists of two steps: (1) querying node-level prototypes from a Meta-Node Bank; (2) reconstructing node embeddings with Hyper-Network \cite{ha2016hypernetworks}. This localized memorizing capability empowers our modularized Meta-Graph Learner to essentially distinguish traffic patterns on different roads over time, which is even generalizable to incident situations. Our contributions are highlighted as follows:
\begin{itemize}
    \item We propose a novel Meta-Graph Learner for spatio-temporal graph (STG) learning, to explicitly disentangles the heterogeneity in space and time.
    \item We present a generic Meta-Graph Convolutional Recurrent Network (MegaCRN), which simply relies on observational data to be robust and adaptive to any traffic situation, from normal to non-stationary.
    \item We validate MegaCRN quantitatively and qualitatively over a group of state-of-the-art models on two benchmarks (METR-LA and PEMS-BAY) and our newly published dataset (EXPY-TKY) that has larger scale and more complex incident situations.
\end{itemize}
\section{Related Work}
\noindent\textbf{Traffic Forecasting.} Traffic forecasting has been taken as a significant research problem in transportation engineering~\cite{huang2014deep,lv2014traffic,ma2015large}. As a canonical case of multivariate time series forecasting~\cite{lai2018modeling}, it also has drawn a lot of attention from machine learning researchers. At the very beginning, statistical models including autoregressive model (AR)~\cite{hamilton1994autoregressive}, vector autoregression (VAR)~\cite{stock2001vector}, autoregressive integrated moving average (ARIMA)~\cite{pan2012utilizing} were applied. Then deep learning methods come to dominate the time series prediction by automatically extracting the non-linear complex features from the data. First, LSTM~\cite{hochreiter1997long} and GRU~\cite{chung2014empirical} demonstrated superior performance in traffic modeling~\cite{ma2015long,lv2018lc,li2018diffusion,zhao2019t,wang2020traffic,bai2020adaptive,ye2021coupled,shang2021discrete,lee2021learning} as well as multivariate time series forecasting~\cite{lai2018modeling,shih2019temporal}. Second, instead of the RNNs, Temporal Convolution~\cite{yu2015multi} and WaveNet~\cite{oord2016wavenet} with long receptive field were also utilized as the core component in~\cite{yu2018spatio,wu2019graph,wu2020connecting,lu2020spatiotemporal,deng2021st} for temporal modeling. Third, motivated by~\cite{vaswani2017attention}, a series of traffic transformers~\cite{zheng2020gman,xu2020spatial} and time series transformers~\cite{li2019enhancing,zhou2021informer,xu2021autoformer} were proposed to do the long sequence time series modeling. Due to the space limitation, we refer you to the recent surveys~\cite{jiang2021dl,jiang2021graph,li2021dynamic} on traffic forecasting with deep learning.


\noindent\textbf{Graph Structure Learning.} Besides the sequence modeling, research efforts have been made to capture the correlations among variables (road links in traffic data) via generic graph structures~\cite{kipf2018neural}.
Early methods either rely on the natural topology of the road network (i.e., binary adjacency graph) or pre-defined graphs in certain metrics (e.g., Euclidean distance)~\cite{li2018diffusion,yu2018spatio}. Then, GW-Net~\cite{wu2019graph} first proposed to use two learnable embedding matrices to automatically build an adaptive graph based on the input traffic data. Following GW-Net~\cite{wu2019graph}, MTGNN~\cite{wu2020connecting} and GTS~\cite{shang2021discrete} further proposed to learn a parameterized k-degree discrete graph, while AGCRN~\cite{bai2020adaptive} introduced node-specific convolution filters according to the node embedding and CCRNN~\cite{ye2021coupled} learned multiple adaptive graphs for multi-layer graph convolution. StemGNN~\cite{cao2020spectral} took the self-attention~\cite{vaswani2017attention} learned from the input as the latent graph. Our work distinguishes itself from these methods by augmenting the spatio-temporal graph learning with memory network (Meta-Node Bank) to discover latent node-level prototypes and construct memory-tailored node embedding.


\section{Problem Definition}
Without loss of generality, we formulate our problem as a multi-step-to-multi-step forecasting task as follows: 
\begin{equation}
    [X_{t-(\alpha-1)}, ..., X_{t}] \xrightarrow[\theta]{ \quad \mathbb{F}(\cdot) \quad}  [X_{t+1}, ..., X_{t+\beta}]
\end{equation}
where $X_i$ $\in$ $\mathbb{R}^{N \times C}$, $N$ is the number of spatial units (i.e., nodes, grids, regions, road links, etc.), and $C$ is the number of the information channel. In our case, $C$ is equal to 1 as we only forecast the traffic speed; the spatial unit is road link. To be simple, we omit $C$ in the rest of our paper. Given previous $\alpha$ steps of observations [$X_{t-(\alpha-1)}$,...,$X_{t-1}$,$X_{t}$], 
we aim to infer the next $\beta$ horizons [$X_{t+1}$,$X_{t+2}$,...,$X_{t+\beta}$] by training a forecasting model $\mathbb{F}$ with parameter $\theta$. 
    
\section{Methodology}
\begin{figure*}[h]
	\centering
	\includegraphics[width=0.9\textwidth]{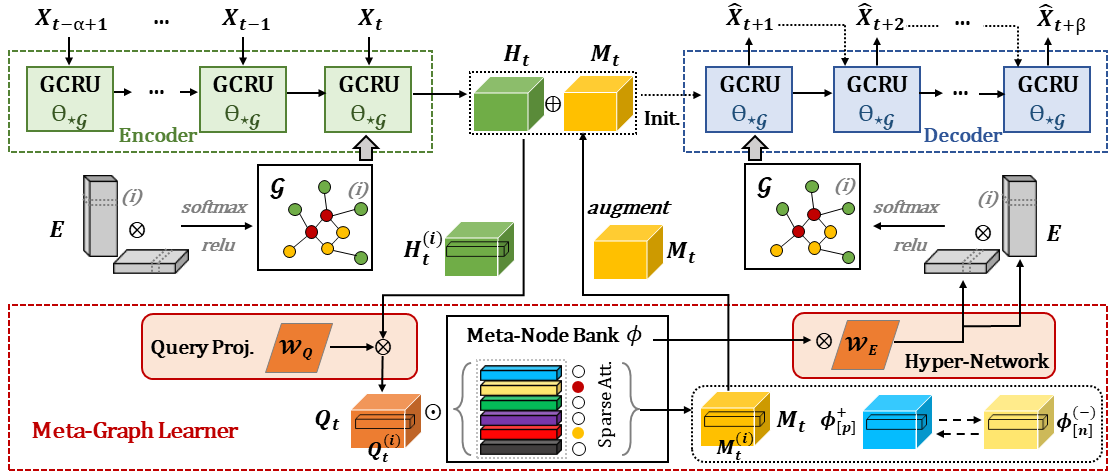}
	\caption{Framework of \textbf{\underline{Me}}ta-\textbf{\underline{G}}r\textbf{\underline{a}}ph \underline{\textbf{C}}onvolutional \textbf{\underline{R}}ecurrent \textbf{\underline{N}}etwork (\textbf{MegaCRN})}
	\label{fig:architecture}
\end{figure*}

In this section, we present a generic framework for spatio-temporal meta-graph learning, namely Meta-Graph Convolutional Recurrent Network (\textbf{MegaCRN}), built upon Graph Convolutional Recurrent Unit (GCRU) Encoder-Decoder and plugin Meta-Graph Learner, as illustrated in Figure \ref{fig:architecture}.

\subsection{Preliminaries}
\noindent\textbf{Graph Convolutional Recurrent Unit.} Motivated by the success of Graph Convolutional Networks (GCNs) as a class in representation learning on graph-structured data (\textit{e.g.} social and road networks), a recent line of research \cite{li2018diffusion, bai2020adaptive, shang2021discrete, ye2021coupled} has explored the possibility of injecting graph convolution operation into recurrent cell (\textit{e.g.} LSTM). The derived Graph Convolutional Recurrent Unit (GCRU) can thereby simultaneously capture spatial dependency, represented by an input graph topology, and temporal dependency in a sequential manner. Without loss of generality, we take the widely adopted definitions of graph convolution operation and Gated Recurrent Unit (GRU) to denote GCRU, as the basic unit for spatio-temporal modeling:
\begin{equation} \label{eq:gcn}
	H = \sigma (X \star_\mathcal{G} \Theta) = \sigma (\mathop{\sum}_{k=0}^K \mathcal{\tilde P}^k X \textit{ } W_k)
\end{equation}
\begin{equation} \label{eq:gcru}
	\begin{cases}
	\begin{aligned}
		u_t & = sigmoid([X_t, \textit{ } H_{t-1}] \star_\mathcal{G} \Theta_{u} + b_{u})      \\
		r_t & = sigmoid([X_t, \textit{ } H_{t-1}] \star_\mathcal{G} \Theta_{r} + b_{r})      \\
		C_t & = tanh([X_t, \textit{ } (r_t \odot H_{t-1})] \star_\mathcal{G} \Theta_{C} + b_{C})      \\
		H_t & = u_t \odot H_{t-1} + (1 - u_t) \odot C_t
	\end{aligned}
	\end{cases}
\end{equation}

In Equation (\ref{eq:gcn}), $X \in \mathbb{R}^{N \times C}$ and $H \in \mathbb{R}^{N \times h}$ denote the input and output of graph convolution operation ($\star_\mathcal{G}$), in which $\Theta$ or $W_K \in \mathbb{R}^{K \times C \times h}$ are the kernel parameters approximated with the Chebyshev polynomials to the order of $K$ \cite{defferrard2016convolutional} and $\sigma$ is an activation function. In Equation (\ref{eq:gcru}), subscripts \textit{u}, \textit{r}, and \textit{C} denote update gate, reset gate, and candidate state in a GCRU cell, in which $\Theta_{\{u,r,C\}} \in \mathbb{R}^{K \times (C+h) \times h}$ denote the gate parameters. Besides observation $X_t$, GCRU requires an auxiliary input $\mathcal{P} \in \mathbb{R}^{N \times N}$ for the topology of graph $\mathcal{G}$.

\noindent\textbf{Graph Structure Learning.} Matrix $\mathcal{P}$ is conventionally defined based on certain metrics (\textit{e.g.} inverse distance, cosine similarity) and empirical laws \cite{yu2018spatio, li2018diffusion, geng2019spatiotemporal}. However, choice of metric can be arbitrary and suboptimal, which motivates a line of research \cite{wu2020connecting, zhang2020spatio, shang2021discrete, bai2020adaptive, ye2021coupled} to integrate Graph Structure Learning (GSL) into spatio-temporal modeling for simultaneous optimization. Here we adopt the canonical formulation \cite{wu2019graph, bai2020adaptive, wang2022event} for spatio-temporal graph learning, namely \textit{adaptive} graph (in Figure \ref{fig:intro}), denoted by:
\begin{equation} \label{eq:adaG}
	\tilde{\mathcal{P}} = softmax(relu(E \textit{ } E^{\top}))
\end{equation}
where $\tilde{\mathcal{P}}$ is derived by random walk normalizing the non-negative part of matrix product of trainable node embedding $E \in \mathbb{R}^{N \times e}$ to its transpose. The other GSL strategy, \textit{momentary} graph \cite{zhang2020spatio} (in Figure \ref{fig:intro}), can be defined in a similar fashion with input signal $X_t$ or hidden state $H_t$. Taking the latter as an example:
\begin{equation} \label{eq:momG}
    \tilde{\mathcal{P}}_t = softmax(relu((H_t * W) \textit{ } (H_t * W)^{\top}))
\end{equation}
where parameter matrix $W \in \mathbb{R}^{h \times e}$ essentially projects $H_t$ to another embedding space. Notably, \textit{momentary} graph has other variants, such as replacing the projection with self-attention operation \cite{cao2020spectral}, but they still follow Equation (\ref{eq:momG}) as a general form.

\subsection{Spatio-Temporal Meta-Graph Learner}
Here we formally describe a new spatio-temporal graph (STG) learning module. The term \textit{meta-graph} is coined to represent the generation of node embedding for graph structure learning, which is different from its definition in heterogeneous information networks (HIN) \cite{zhao2017meta, ding2021diffmg}. According to the definition in Equation (\ref{eq:adaG}) and (\ref{eq:momG}), \textit{adaptive} graph relies on parameterized node embedding $E$ alone, while \textit{momentary} graph is in fact input-conditioned (either projecting $X_t$ or $H_t$ with parameter $W$). Apparently, this generation process determines the properties of the derived graphs, as the former is time-invariant but the latter is sensitive to input signals. This motivates us to further enhance the node embeddings for STG generation, as the real-world networks are more complex, manifesting spatio-temporal heterogeneity and non-stationarity.

We are inspired by a line of research in memory networks, which aims to memorize typical features in seen samples for further pattern matching. This technique has been largely employed on computer vision tasks, such as few-shot learning \cite{vinyals2016matching, santoro2016meta} and anomaly detection \cite{gong2019memorizing, park2020learning}. In our case, we would like inject the memorizing and distinguishing capabilities into spatio-temporal graph learning. We thereby leverage the idea of memory networks and build a \textbf{Meta-Node Bank} $\Phi \in \mathbb{R}^{\phi \times d}$. Here $\phi$ and $d$ denote the number of memory items and the dimension of each item, respectively. We further define the main functions of this memory bank as follows:
\begin{equation} \label{eq:mem-query}
	Q_t^{(i)} = H_t^{(i)} * W_Q + b_Q
\end{equation}
\vspace{-0.3cm}
\begin{equation} \label{eq:mem-recon}
	\begin{cases}
	\begin{aligned}
		a_j^{(i)} & = \frac{\exp{(Q_t^{(i)} * \Phi^{\top}[j])}}{\sum_{j=1}^{\phi} \exp{(Q_t^{(i)} * \Phi^{\top}[j])}}     \\
		M_t^{(i)} & = \sum_{j=1}^{\phi} a_j^{(i)} * \Phi[j]
	\end{aligned}
	\end{cases}
\end{equation}
where we use superscript $(i)$ as row index. For instance, $H_t^{(i)} \in \mathbb{R}^h$ represents $i$-th node vector in $H_t \in \mathbb{R}^{N \times h}$. Equation (\ref{eq:mem-query}) denotes a fully connected (FC parameter $W_Q \in \mathbb{R}^{h \times d}$) layer to project hidden state $H_t^{(i)}$ to a localized query $Q_t^{(i)} \in \mathbb{R}^d$. Equation (\ref{eq:mem-recon}) denotes the memory reading operation by matching $Q_t^{(i)}$ with each memory $\Phi[j]$ to calculate a scalar $a_j$, which physically represents the similarity between vector $Q_t^{(i)}$ and memory item $\Phi[j]$. A meta-node vector $M_t^{(i)} \in \mathbb{R}^d$ can be further recovered as a combination of memory items. Here a common practice is to utilize the reconstructed representation $M_t \in \mathbb{R}^{N \times d}$ to augment the encoded hidden representation $H_t$, denoted by $H'_t = [H_t, M_t] \in \mathbb{R}^{N \times (h+d)}$ ($[\cdot]$ denotes a concatenation operation) \cite{yao2019learning, lee2021learning}. We further leverage a Hyper-Network \cite{ha2016hypernetworks} that essentially puts generation of GSL node embeddings conditioned on Meta-Node Bank. This memory-augmented node embedding generation can be formulated as:
\begin{equation} \label{eq:hyper-node}
	\begin{cases}
	\begin{aligned}
		E' & = \textit{NN}_H(\Phi)     \\
		\tilde{\mathcal{P}'} & = softmax(relu(E^{'} \textit{ } E^{'\top}))
	\end{aligned}
	\end{cases}
\end{equation}
where $\textit{NN}_H$ denotes a Hyper-Network. Without loss of generality, we implement it with one FC layer (parameter $W_E \in \mathbb{R}^{d \times e}$). Then, \textit{meta-graph} $\tilde{\mathcal{P}'}$ can be constructed, as an alternative to \textit{adaptive} and \textit{momentary} graphs (defined in Equation (\ref{eq:adaG}) an (\ref{eq:momG})) to feed back to GCRU encoder-decoder. 

\subsection{Meta-Graph Convolutional Recurrent Network}
With Meta-Graph Learner as described, we present the proposed Meta-Graph Convolutional Recurrent Network (\textbf{MegaCRN}) as a generic framework for spatial-temporal modeling. MegaCRN learns node-level prototypes of traffic patterns in Meta-Node Bank for updating  the auxiliary graph adaptively based on the observed situation. To further enhance its distinguishing power for diverse scenarios on different roads over time, we regulate the memory parameters with two constraints \cite{gong2019memorizing, park2020learning}, including a contrastive loss $\mathcal{L}_{1}$ and a consistency loss $\mathcal{L}_{1}$, denoted by:
\begin{equation} \label{eq:mem-loss}
	    \begin{aligned}
         \mathcal{L}_{1} & = \mathop{\sum}_t^T \mathop{\sum}_i^N \max\{||Q_t^{(i)} - \Phi[p]||^2 - ||Q_t^{(i)} - \Phi[n]||^2 + \lambda, 0\}\\
		\mathcal{L}_{2} & = \mathop{\sum}_t^T \mathop{\sum}_i^N ||Q_t^{(i)} - \Phi[p]||^2 
	    \end{aligned}
\end{equation}
where $T$ denotes the total number of sequences (\textit{i.e.} samples) in the training set and $p, n$ denote the top two indices of memory items by ranking $a_j^{(i)}$ in Equation~\ref{eq:mem-recon} given localized query $Q_t^{(i)}$. By implementing these two constraints, we treat $Q_{t}^{(i)}$ as anchor, its most similar prototype $\Phi[p]$ as positive sample, and the second similar prototype $\Phi[n]$ as negative sample ($\lambda$ denotes the margin between the positive and negative pairs). Here the idea is to keep memory items as compact as possible, at the same time as dissimilar as possible. These two constraints guide memory $\Phi$ to distinguish different spatio-temporal patterns on node-level. In practice, we find adding them into the objective criterion (\textit{i.e.} MAE) facilitates the convergence of training (with balancing factors $\kappa_1, \kappa_2$):
\begin{equation}    \label{eq:task-loss}
    \mathcal{L}_{task} = \mathop{\sum}_t^T \mathop{\sum}_\rho^\beta |\hat{X}_{t+\rho} - X_{t+\rho}| + \kappa_1 \mathcal{L}_{1} + \kappa_2 \mathcal{L}_{2}
\end{equation}

\section{Experiment}\label{sec:experiment}
\begin{table}[h]
  \footnotesize
  \centering
  \caption{Summary of Datasets}
  \label{tab:datasummary}
  \begin{tabular}{llll}
    \hline
	Dataset & METR-LA & PEMS-BAY & EXPY-TKY\\
    \hline
    Start Time & 2012/3/1 & 2017/1/1 & 2021/10/1\\
    End Time & 2012/6/30 & 2017/5/31 & 2021/12/31\\
    Time Interval & 5 minutes & 5 minutes & 10 minutes\\
    \#Timesteps & 34,272 & 52,116 & 13,248 \\
    \#Spatial Units & 207 sensors & 325 sensors & 1,843 road links\\
    \hline
\end{tabular}
\end{table}
\begin{table*}[h]
    \footnotesize
    \centering
    \caption{Forecasting Performance}
	\label{tab:benchmark}
	\resizebox{!}{73mm}{
	\begin{tabular*}{17.6cm}{@{\extracolsep{\fill}}cccccccccc}
		\hline
		\multirow{2}{*}{\textbf{METR-LA}} & \multicolumn{3}{c}{15min / horizon 3} &
		\multicolumn{3}{c}{30min / horizon 6} &
		\multicolumn{3}{c}{60min / horizon 12} \\
		\cline{2-4} \cline{5-7} \cline{8-10}
		\multicolumn{1}{l}{} & 
		\multicolumn{1}{c}{MAE} & 
		\multicolumn{1}{c}{RMSE} &
		\multicolumn{1}{c}{MAPE} &
		\multicolumn{1}{c}{MAE} & 
		\multicolumn{1}{c}{RMSE} &
		\multicolumn{1}{c}{MAPE} &
		\multicolumn{1}{c}{MAE} & 
		\multicolumn{1}{c}{RMSE} &
		\multicolumn{1}{c}{MAPE} \\
		\hline
		HA\cite{li2018diffusion} & 4.16 & 7.80 & 13.00\% & 4.16 & 7.80 & 13.00\% & 4.16 & 7.80 & 13.00\% \\
		STGCN\cite{yu2018spatio} & 2.88 & 5.74 & 7.62\% & 3.47 & 7.24 & 9.57\% & 4.59 & 9.40 & 12.70\% \\
		DCRNN\cite{li2018diffusion} & 2.77 & 5.38 & 7.30\% & 3.15 & 6.45 & 8.80\% & 3.60 & 7.59 & 10.50\% \\
		GW-Net\cite{wu2019graph} & 2.69 & 5.15 & 6.90\% & 3.07 & 6.22 & 8.37\% & 3.53 & 7.37 & 10.01\% \\
		STTN\cite{xu2020spatial} & 2.79 & 5.48 & 7.19\% & 3.16 & 6.50 & 8.53\% & 3.60 & 7.60 & 10.16\% \\
        GMAN\cite{zheng2020gman}$\star$ & 2.80 & 5.55 & 7.41\% & 3.12 & 6.49 & 8.73\% & 3.44 & 7.35 & 10.07\% \\ 
		MTGNN\cite{wu2020connecting} & 2.69 & 5.18 & 6.86\% & 3.05 & 6.17 & 8.19\% & 3.49 & 7.23 & 9.87\% \\
		StemGNN\cite{cao2020spectral}$\dagger$ & 2.56 & 5.06 & 6.46\% & 3.01 & 6.03 & 8.23\% & 3.43 & 7.23 & 9.85\% \\
		AGCRN\cite{bai2020adaptive} & 2.86 & 5.55 & 7.55\% & 3.25 & 6.57 & 8.99\% & 3.68 & 7.56 & 10.46\% \\
        CCRNN\cite{ye2021coupled} & 2.85 & 5.54 & 7.50\% & 3.24 & 6.54 & 8.90\% & 3.73 & 7.65 & 10.59\% \\
        GTS\cite{shang2021discrete}$\star$ & 2.65 & 5.20 & 6.80\% & 3.05 & 6.22 & 8.28\% & 3.47 & 7.29 & 9.83\% \\
		PM-MemNet\cite{lee2021learning} & 2.65 & 5.29 & 7.01\% & 3.03 & 6.29 & 8.42\% & 3.46 & 7.29 & 9.97\% \\
          \textbf{MegaCRN (Ours)} & \textbf{2.52} & \textbf{4.94} & \textbf{6.44\%} & \textbf{2.93} & \textbf{6.06} & \textbf{7.96\%} & \textbf{3.38} & \textbf{7.23} & \textbf{9.72\%} \\
		\hline
		\multirow{2}{*}{\textbf{PEMS-BAY}} & \multicolumn{3}{c}{15min / horizon 3} &
		\multicolumn{3}{c}{30min / horizon 6} &
		\multicolumn{3}{c}{60min / horizon 12} \\
		\cline{2-4} \cline{5-7} \cline{8-10}
		\multicolumn{1}{l}{} & 
		\multicolumn{1}{c}{MAE} & 
		\multicolumn{1}{c}{RMSE} &
		\multicolumn{1}{c}{MAPE} &
		\multicolumn{1}{c}{MAE} & 
		\multicolumn{1}{c}{RMSE} &
		\multicolumn{1}{c}{MAPE} &
		\multicolumn{1}{c}{MAE} & 
		\multicolumn{1}{c}{RMSE} &
		\multicolumn{1}{c}{MAPE} \\
		\hline
		HA\cite{li2018diffusion} & 2.88 & 5.59 & 6.80\% & 2.88 & 5.59 & 6.80\% & 2.88 & 5.59 & 6.80\% \\
		STGCN\cite{yu2018spatio} & 1.36 & 2.96 & 2.90\% & 1.81 & 4.27 & 4.17\% & 2.49 & 5.69 & 5.79\% \\
		DCRNN\cite{li2018diffusion} & 1.38 & 2.95 & 2.90\% & 1.74 & 3.97 & 3.90\% & 2.07 & 4.74 & 4.90\% \\
		GW-Net\cite{wu2019graph} & 1.30 & 2.74 & 2.73\% & 1.63 & 3.70 & 3.67\% & 1.95 & 4.52 & 4.63\% \\
		STTN\cite{xu2020spatial} & 1.36 & 2.87 & 2.89\% & 1.67 & 3.79 & 3.78\% & 1.95 & 4.50 & 4.58\% \\
        GMAN\cite{zheng2020gman}$\star$ & 1.35 & 2.90 & 2.87\% & 1.65 & 3.82 & 3.74\% & 1.92 & 4.49 & 4.52\% \\ 
	MTGNN\cite{wu2020connecting} & 1.32 & 2.79 & 2.77\% & 1.65 & 3.74 & 3.69\% & 1.94 & 4.49 & 4.53\% \\
		StemGNN\cite{cao2020spectral}$\dagger$ & 1.23 & 2.48 & 2.63\% & \multicolumn{3}{c}{N/A from \cite{cao2020spectral}} & \multicolumn{3}{c}{N/A from \cite{cao2020spectral}} \\
		AGCRN\cite{bai2020adaptive} & 1.36 & 2.88 & 2.93\% & 1.69 & 3.87 & 3.86\% & 1.98 & 4.59 & 4.63\% \\
        CCRNN\cite{ye2021coupled} & 1.38 & 2.90 & 2.90\% & 1.74 & 3.87 & 3.90\% & 2.07 & 4.65 & 4.87\% \\
        GTS\cite{shang2021discrete}$\star$ & 1.34 & 2.84 & 2.83\% & 1.67 & 3.83 & 3.79\% & 1.98 & 4.56 & 4.59\% \\
        PM-MemNet\cite{lee2021learning}$\star$ & 1.34 & 2.82 & 2.81\% & 1.65 & 3.76 & 3.71\% & 1.95 & 4.49 & 4.54\% \\
    	\textbf{MegaCRN (Ours)} & \textbf{1.28} & \textbf{2.72} & \textbf{2.67\%} & \textbf{1.60} & \textbf{3.68} & \textbf{3.57\%} & \textbf{1.88} & \textbf{4.42} & \textbf{4.41\%} \\
		\hline
		\multirow{2}{*}{\textbf{EXPY-TKY}} & \multicolumn{3}{c}{10min / horizon 1} &
		\multicolumn{3}{c}{30min / horizon 3} &
		\multicolumn{3}{c}{60min / horizon 6} \\
		\cline{2-4} \cline{5-7} \cline{8-10}
		\multicolumn{1}{l}{} & 
		\multicolumn{1}{c}{MAE} & 
		\multicolumn{1}{c}{RMSE} &
		\multicolumn{1}{c}{MAPE} &
		\multicolumn{1}{c}{MAE} & 
		\multicolumn{1}{c}{RMSE} &
		\multicolumn{1}{c}{MAPE} &
		\multicolumn{1}{c}{MAE} & 
		\multicolumn{1}{c}{RMSE} &
		\multicolumn{1}{c}{MAPE} \\
		\hline
		HA\cite{li2018diffusion} & 7.63 & 11.96 & 31.26\% & 7.63 & 11.96 & 31.25\% & 7.63 & 11.96 & 31.24\% \\
		STGCN\cite{yu2018spatio} & 6.09 & 9.60 & 24.84\% & 6.91 & 10.99 & 30.24\% & 8.41 & 12.70 & 32.90\%\\
		DCRNN\cite{li2018diffusion} & 6.04 & 9.44 & 25.54\% & 6.85 & 10.87 & 31.02\% & 7.45 & 11.86 & 34.61\%\\
		GW-Net\cite{wu2019graph} & 5.91 & 9.30 & 25.22\% & 6.59 & 10.54 & 29.78\% & 6.89 & 11.07 & 31.71\% \\
		STTN\cite{xu2020spatial} & 5.90 & 9.27 & 25.67\% & 6.53 & 10.40 & 29.82\% & 6.99 & 11.23 & 32.52\%\\
		GMAN\cite{zheng2020gman} & 6.09 & 9.49 & 26.52\% & 6.64 & 10.55 & 30.19\% & 7.05 & 11.28 & 32.91\%\\
		MTGNN\cite{wu2020connecting} & 5.86 & 9.26 & 24.80\% & 6.49 & 10.44 & 29.23\% & \textbf{6.81} & \textbf{11.01} & 31.39\% \\
		StemGNN\cite{cao2020spectral}$\dagger$ & 6.08 & 9.46 & 25.87\% & 6.85 & 10.80 & 31.25\% & 7.46 & 11.88 & 35.31\%\\
		AGCRN\cite{bai2020adaptive} & 5.99 & 9.38 & 25.71\% & 6.64 & 10.63 & 29.81\% & 6.99 & 11.29 & 32.13\% \\
		CCRNN\cite{ye2021coupled} & 5.90 & 9.29 & 24.53\% & 6.68 & 10.77 & 29.93\% & 7.11 & 11.56 & 32.56\%\\
		GTS\cite{shang2021discrete} & - & - & - & - & - & - & - & - & - \\
		PM-MemNet\cite{lee2021learning} & 5.94 & 9.25 & 25.10\% & 6.52 & 10.42 & 29.00\% & 6.87 & 11.14 & 31.22\% \\
        \textbf{MegaCRN (Ours)} & \textbf{5.81} & \textbf{9.20} & \textbf{24.49\%} & \textbf{6.44} & \textbf{10.33} & \textbf{28.92\%} & \underline{6.83} & \underline{11.04} & \textbf{31.02\%} \\
		\hline
	\end{tabular*}
	}
	\centering
\end{table*}

\subsection{Datasets and Settings}\label{sec:experiment-setup}
\noindent\textbf{Datasets.} We first evaluate our model by using two standard benchmark datasets from \cite{li2018diffusion}: \textbf{METR-LA} and \textbf{PEMS-BAY}. They contain the traffic speed data from 207 sensors in Los Angeles and 325 sensors in Bay Area respectively. For the two benchmarks, we follow the tradition~\cite{li2018diffusion,wu2019graph,shang2021discrete,lee2021learning} by splitting the datasets in chronological order with 70\% for training, 10\% for validation, and 20\% for testing (namely 7:1:2). Besides, in this study, we publish a new traffic dataset called \textbf{EXPY-TKY}, that contains the traffic speed information and the corresponding traffic incident information in 10-minute interval for 1843 expressway road links in Tokyo over three months (2021/10$\sim$2021/12). We use the first two months (Oct. 2021 and Nov. 2021) as the training and validation dataset, and the last one month (Dec. 2021) as the testing dataset. The specific spatio-temporal information of our datasets are summarized in Table \ref{tab:datasummary}. 

\noindent\textbf{Settings.}
For EXPY-TKY, the Encoder and Decoder of our model consist of 1 RNN-layer respectively, where the number of hidden states is 32. We reserve 10 prototypes (i.e., meta-nodes) in the memory, each of which is a 32-dimension learnable vector. For METR-LA and PEMSBAY, each RNN layer in Encoder and Decoder has 64 units and the memory bank has 20 meta-nodes with 64-dimension. The observation step $\alpha$ and prediction horizon $\beta$ are both set to 12 on METR-LA and PEMS-BAY, while $\alpha$/$\beta$ are both set to 6 on EXPY-TKY. Such settings can give us 1-hour lead time forecasting, which follow the tradition in previous literatures~\cite{yu2018spatio,li2018diffusion,wu2019graph,bai2020adaptive,shang2021discrete}. Adam was used as the optimizer, where the learning rate was set to 0.01 and the batch size was set to 64. The optimizer would either be early-stopped if the validation error was converged within 20 epochs or be stopped after 200 epochs. L1 Loss is used as the loss function. Root Mean Square Error (RMSE), Mean Absolute Error (MAE), and Mean Absolute Percentage Error (MAPE) are used as metrics. All experiments were performed with four GeForce RTX 3090 GPUs. 

\subsection{Quantitative Evaluation}\label{sec:experiment-overall}

We compare our model with the following baselines: 1) Historical Average (HA) averaged values of the same time slot from historical days~\cite{li2018diffusion}; 2) STGCN~\cite{yu2018spatio}, 3) DCRNN~\cite{li2018diffusion}, and 4) GW-Net~\cite{wu2019graph}, the most representative deep models for traffic forecasting, respectively embed spectral~\cite{yu2018spatio} or diffusion graph convolution~\cite{li2018diffusion,wu2019graph} into temporal convolution (i.e., TCN or WaveNet)\cite{yu2018spatio,wu2019graph} or recurrent unit (e.g., GRU)\cite{li2018diffusion}; 5) STTN~\cite{xu2020spatial} and 6) GMAN~\cite{zheng2020gman} are two Transformer-based SOTAs; 7) MTGNN~\cite{wu2020connecting} is an extended version of GW-Net that extends the adaptive graph leaning part;
8) StemGNN~\cite{cao2020spectral} first learns a latent graph via self-attention and performs the spatiotemporal modeling in spectral domain; 9) AGCRN~\cite{bai2020adaptive} adaptively learns node-specific parameters for graph convolution; 10) CCRNN~\cite{ye2021coupled} learns multiple parameterized matrices for multiple layers of graph convolution; 11) GTS~\cite{shang2021discrete} learns each link's (edge's) probability based on each variable's (node's) long historical data; 12) PM-MemNet~\cite{lee2021learning} also utilizes memory networks for traffic pattern matching. 9)$\sim$12) are built based upon GCRN~\cite{seo2018structured,li2018diffusion}. 


\noindent\textbf{Overall Comparison.} Most of the baselines' results on the benchmarks are reported from their original papers. However, due to the reproducibility problem ($\star$ in Table~\ref{tab:benchmark}), we report the results of GMAN, GTS, and PM-MemNet either from our own experiments or other literature (e.g., GMAN on METR-LA~\cite{shao2022pre}). Through Table~\ref{tab:benchmark}, we can find our model outperformed the state-of-the-arts in almost all cases (dataset/horizon/metric). Among the SOTAs, GW-Net\cite{wu2019graph} and MTGNN~\cite{wu2020connecting} marked relatively good performances thanks to the WaveNet backbone. CCRNN~\cite{ye2021coupled} delivered better performance on our dataset than the benchmarks. Because it requires the 0-1 adjacency matrix of the road network to get a good initialization for the learnable graphs, which is not available in the benchmark datasets. GMAN~\cite{zheng2020gman} and StemGNN~\cite{cao2020spectral} gave a worse performance on our dataset, because the number of nodes $N$ in ours is around 6$\sim$9 times larger than the benchmarks and the self-attention in them struggled to work on such a large scale. GTS~\cite{shang2021discrete} could not even be applicable on our dataset, because it requires to parameterize an $\mathbb{R}^{N^2\times N}$ matrix ($N^2$ edges and $N$ nodes) for edge generation based on each node's features. StemGNN$\dagger$ is considered to have a \textit{data leakage} problem as it averages all the graphs in minibatch for both training and testing (sample interaction). 


\begin{table}[h]
    \footnotesize
    \centering
	\caption{Ablation Test across All Horizons}
	\label{tab:ablationstudy}
	\setlength{\tabcolsep}{0.8mm}{
	\begin{tabular*}{8.2cm}{@{\extracolsep{\fill}}lccc}
		\hline
		\multirow{2}{*}{Ablation} & METR-LA & PEMS-BAY & EXPY-TKY \\
		\cline{2-4}
		\multicolumn{1}{l}{} & 
		\multicolumn{1}{c}{MAE / RMSE} & 
		\multicolumn{1}{c}{MAE / RMSE} & 
		\multicolumn{1}{c}{MAE / RMSE} \\
		\hline
		Adaptive & 3.01 / 6.25 & 1.61 / 3.73 & 6.79 / 10.76\\
            Momentary & 2.96 / 6.16 & 1.62 / 3.75 & 6.68 / 10.59 \\		
            Memory & 2.97 / 6.21 & 1.60 / 3.70 & 6.55 / 10.48 \\
		\hline
		\textbf{MegaCRN} & \textbf{2.89} / \textbf{6.02} & \textbf{1.54} / \textbf{3.59}  & \textbf{6.44} / \textbf{10.35} \\
		\hline
	\end{tabular*}}
\end{table}
\noindent\textbf{Ablation Study.} To evaluate the actual performance of each component of our model, we create a series of variants as follows: (1) \textbf{Adaptive GCRN}. It only keeps the GCRN encoder-decoder of MegaCRN and lets the encoder and decoder share a same \textit{adaptive} graph, similar graph structure learning mechanism to GW-Net~\cite{wu2019graph}, MTGCNN~\cite{wu2020connecting}, and AGCRN~\cite{bai2020adaptive}; 
(2) \textbf{Memory GCRN}. It excludes the Hyper-Network from MegaCRN and just uses a Memory Network (i.e., same as the Meta-Node Bank) to get an augmented hidden states $M_t$ (from the encoder) for the decoder, which shares the same \textit{adaptive} graph $\mathcal{G}$ with the encoder.
(3) \textbf{Momentary GCRN}. It excludes the Meta-Node Bank from MegaCRN and directly uses a Hyper-Network (i.e., FC layer) to take the encoder's hidden states $H_t$ to generate a \textit{momentary} graph for the decoder.  Through Table \ref{tab:ablationstudy}, we can see that compared to Momentary GCRN, Memory GCRN brings a higher performance gain to Adaptive GCRN. Because Momentary GCRN is obtained from $H_t$, it has to learn a separate graph for each sample in minibatch, which is non-trivial. All these demonstrate that MegaCRN is a complete and indivisible set.

\begin{figure}[h]
	\centering
	\includegraphics[width=0.9\linewidth]{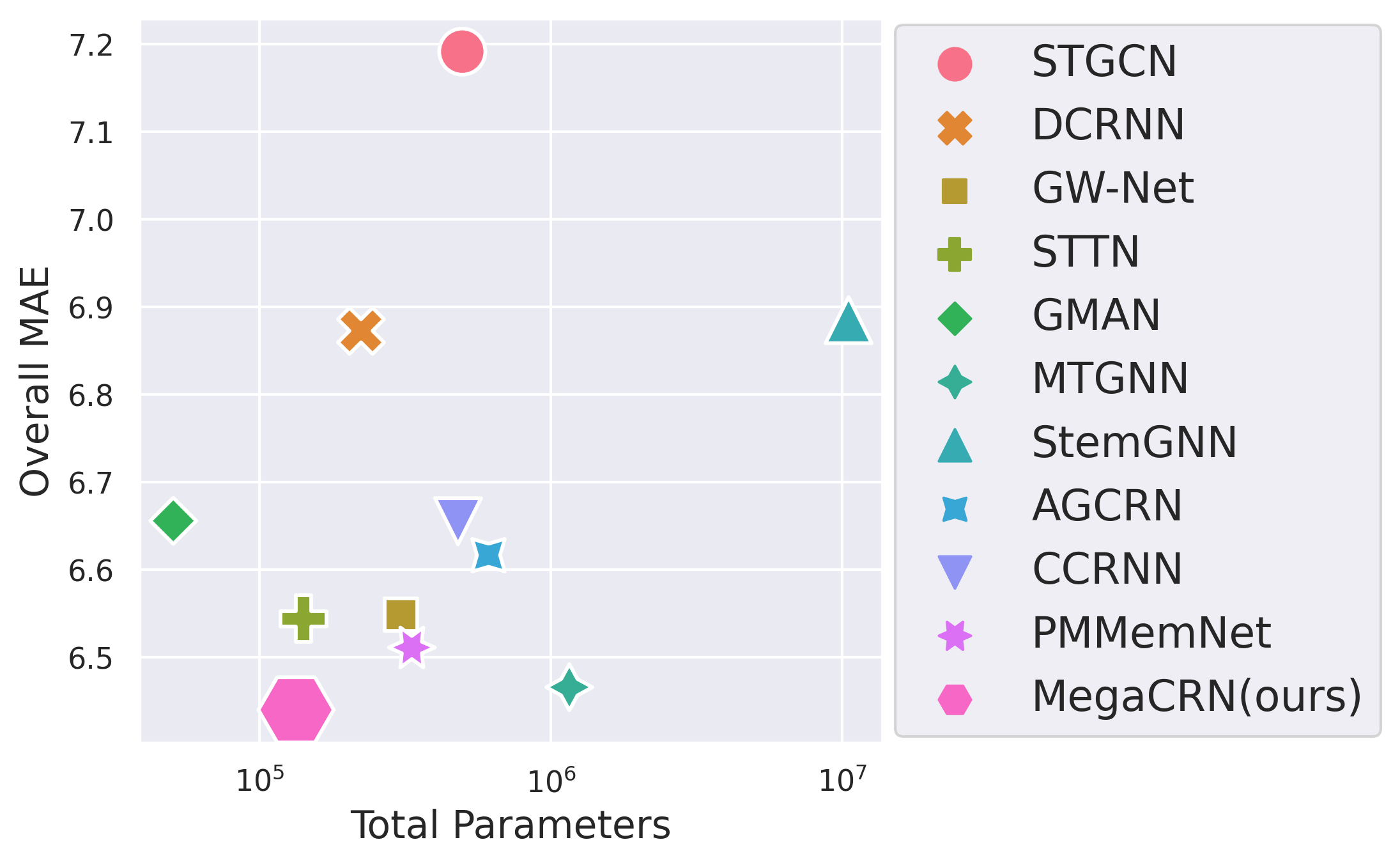}
	\caption{Efficiency Evaluation}
	\label{fig:efficiency}
\end{figure}
\begin{figure*}[h]
	\centering
	\includegraphics[width=0.98\textwidth]{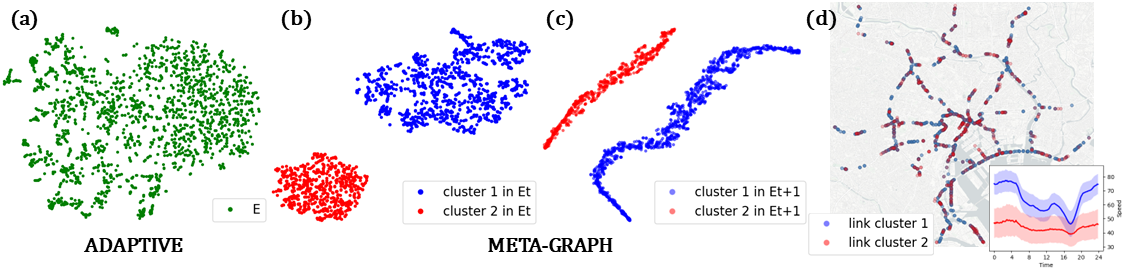}
	\caption{Spatio-Temporal Disentangling Effect of Meta-Graph Learning}
	\label{fig:case-distangle}
\end{figure*}
\begin{figure*}[h]
	\centering
	\includegraphics[width=0.98\textwidth]{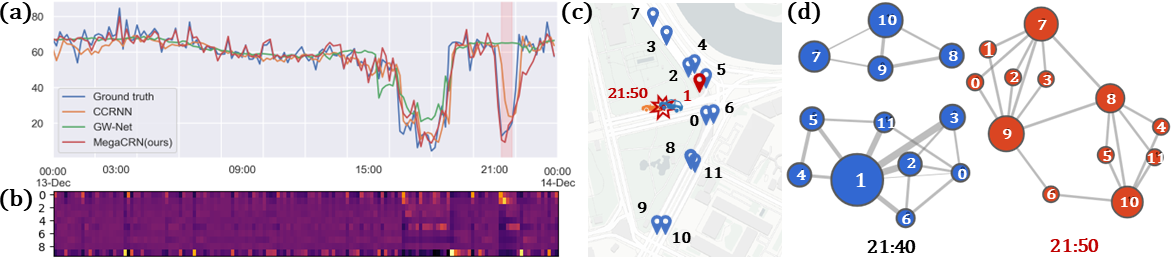}
	\caption{Incident Awareness of MegaCRN}
	\label{fig:case-incident}
\end{figure*}

\noindent\textbf{Efficiency Study.} We also evaluate the efficiency of our model by comparing with the-state-of-the-arts. Here we just report the results on EXPY-TKY, because the spatial domain of our data is 5$\sim$9 times larger than the benchmarks. A scatter plot is shown as Figure \ref{fig:efficiency}, where the x-axis of is the total number of parameters and the y-axis is the overall MAE. We can see that our model has the second-fewest parameters (merely 133,597) but the smallest overall MAE. For a large-scale dataset like EXPY-TKY, our model could be very memory-efficient. In contrast, some models, especially Transformer-based models including GMAN~\cite{zheng2020gman} and STTN~\cite{xu2020spatial}, are very memory/time-consuming due to the dot-product operation on big tensor. Although our model tends to need more epochs to converge, each round of training could be finished in very little time. To sum up, our model can achieve the state-of-the-art precision while keeping comparatively efficient.
\subsection{Qualitative Evaluation} \label{sec:experiment-casestudy}
\textbf{Spatio-Temporal Disentanglement.} We qualitatively evaluate the quality of node embeddings by visualizing them in a low-dimensional space with t-SNE. Compared with \textit{adaptive} GSL illustrated as Figure \ref{fig:case-distangle}(a), \textit{meta-graph} can automatically learn to cluster nodes (i.e., road links) as shown in Figure \ref{fig:case-distangle}(b)(c). Interestingly, as time evolves from t to t+1, this clustering effect persists but the cluster shape changes, which confirms the spatio-temporal disentangling capability as well as the time-adaptability of our approach. In addition, we map out the physical locations of the road links in discovered clusters with different colors (cluster 1 in blue, cluster 2 in red) in Figure \ref{fig:case-distangle}(d). We observe a strong correlation between the spatial distribution of cluster 2 (in red) and interchanges/toll gates. From the daily averaged time series plot in the bottom of Figure \ref{fig:case-distangle}(d), we can clearly tell the inter-cluster difference. While road links in cluster 1 (in blue) share a strong rush hour pattern, the other cluster (in red) has a lower speed on average but higher variations, which is characterized by large amount of speed change near interchanges/toll gates. These observations validate the power of Meta-Graph Learner to explicitly distinguish spatio-temporal heterogeneity. 

\noindent\textbf{Incident-Awareness.} We qualitatively study the robustness of MegaCRN to various traffic situations in Figure \ref{fig:case-incident}. Here we select an incident case that occurred at 21:50 at road link 1, marked in red in Figure \ref{fig:case-incident}(a), on December 13th, 2021. In terms of the prediction results (in 60-minute lead time), compared with two baselines, GW-Net~\cite{wu2019graph} and CCRNN~\cite{ye2021coupled}, our model can not only better capture normal fluctuations, but adapt to more complex situations including rush hour and traffic accident (in shaded red). Such sudden disturbance inevitably results in delay or failure of detection for other models. From the visualization of memory query weight in Figure \ref{fig:case-incident}(b), we can tell that the pattern querying to the Meta-Node Bank is different between normal situations and rush hour or incident case. This observation confirms the distinguishing power and generalizability to diverse traffic scenarios. We further visualize the learned local \textit{meta-graph} as Figure \ref{fig:case-incident}(d), in which thicker line represents higher edge weight and bigger node size means larger weighted outdegree. Intuitively, we can find the \textit{meta-graph} is changing with time. At 21:40 before the accident happened, node 1 (road link 1) held the biggest impact in the local \textit{meta-graph} as road link 1 lies right at the center of the large road intersection. Then at 21:50 after the accident happened, the impact of node 1 dropped significantly and the graph became dominated by road link 7, 8, 9, and 10 that formed a separated subgraph at 21:40. This case study verifies the superior adaptability of our approach. 
\section{Conclusion}\label{sec:conclusion}
In this study, we propose Meta-Graph Convolutional Recurrent Network (MegaCRN) along with a novel spatio-temporal graph structure learning mechanism for traffic forecasting. Besides two benchmarks, METR-LA and PEMS-BAY, we further generate a brand-new traffic dataset called EXPY-TKY from large-scale car GPS records and collect the corresponding traffic incident information. Our model outperformed the state-of-the-arts to a large degree on all three datasets. Through a series of visualizations, it also demonstrated the capability to disentangle the time and nodes with different patterns as well as the adaptability to incident situations. We will further generalize our model for Multivariate Time Series forecasting tasks in the future.


\section{Acknowledgments}
This work was supported by Toyota Motor Corporation and JSPS KAKENHI Grant Number JP20K19859.

\bibliography{aaai23}
\end{document}


\maketitle

\section{Dataset Details}

First, we introduce two standard benchmark traffic datasets, originally published by \cite{li2018diffusion}: \textbf{METR-LA}\cite{zhao2019t,li2018diffusion,wang2020traffic,wu2019graph,zhang2018gaan,chen2020multi,pan2019urban} and \textbf{PEMS-BAY}\cite{li2018diffusion,lai2018modeling,wu2019graph,wang2020traffic,zheng2020gman}. \textbf{METR-LA} is the traffic speed data collected from 207 sensors in Los Angeles County. \textbf{PEMS-BAY} contains 325 traffic sensors in Bay Area, which is collected from Caltrans Performance Measurement System (PEMS) \footnote{\url{http://pems.dot.ca.gov/}}. Please kindly refer to \cite{li2018diffusion} for more details about these two benchmarks. 

\begin{figure}[h]
	\centering
	\includegraphics[width=1.0\linewidth]{./figure/shuto-expy-horizontal.png}
	\caption{Traffic speed on 8 sampled road links in Shuto-Expy dataset.}
	\label{fig:shuto-expy}
\end{figure}

Next, in our study, we publish a brand-new real-world traffic speed dataset called \textbf{Shuto-Expy}. Our dataset covers the traffic speed for 1843 Shuto-Expressway\footnote{\url{https://en.wikipedia.org/wiki/Shuto_Expressway}} road links in Tokyo (longitude $\in$ [139.58, 139.92], latitude $\in$ [35.54, 35.82]) spanned over three months (2021/10$\sim$2021/12). Compared to benchmark datasets, ours is generated based on large-scale GPS records of Toyota connected cars. The original dataset spans over 240 million daily records for more than 50,000 cars over 550,000 road links in the Greater Tokyo Area. The GPS records, where each record contains $timestamp$, $longitude$, $latitude$, $nearest linkid$, $speed$ for each car ID, are collected every 0.1 second. Such a large number of cars and a high sampling rate make our dataset well reflect the real-world traffic situation in one of the biggest metropolitan areas. Furthermore, the dataset also includes a flag about abnormal incidents like accident, construction, obstacle, etc. This range of incident information is generally not available in other public traffic datasets and makes our dataset quite unique and comprehensive. We calculate the average speed for each of the above 1843 links at 10-minute time intervals and also collect the corresponding traffic incident data for each link and time interval. The published data consists of four columns: \textit{timestamp}, \textit{linkid}, \textit{average speed}, \textit{incident flag}. A speed chart for 8 sample links (2 each from inner/outer rings and inner/outer spokes) during an entire day of 2021/12/1 has been plotted in Figure~\ref{fig:shuto-expy} along with their spatial location. Figure~\ref{fig:SpeedHistogram} displays two histograms of speed data for over a day (2021/12/1), and over a month (Dec. 2021), respectively, for a better understanding of how the data is distributed over the period of time. 

\begin{figure}[h]
	\centering	\includegraphics[width=1\linewidth]{./figure/SpeedHistogram.png}	\caption{Histograms of average speed of road links spanned over a day (2021/12/1), and a month (Dec. 2021), respectively.}
	\label{fig:SpeedHistogram}
\end{figure}



\noindent\textbf{Ethics and License.} The raw GPS trajectory data of each car can be only collected under user's consent. The consent is acquired when the customer joins the membership for Toyota Connected Service\footnote{\url{https://www.toyota.com/connected-services/}}.
Since our data just contains aggregated values for traffic speed without any personally identifiable information or objectionable content, it causes no privacy issue to any individual customer. The dataset will be officially published under the MIT License after/if this paper is accepted.

\begin{table*}[h]
    \footnotesize
	\centering
	\caption{Error Bars of Our Model}
	\label{tab:errorbars}
	\begin{tabular*}{16.3cm}{@{\extracolsep{\fill}}lccccccccc}
		\toprule
		\multirow{2}{*}{\textbf{METR-LA}} & 15min / horizon 3 & 30min / horizon 6 & 60min / horizon 12 \\
		 & MAE / RMSE / MAPE(\%) & MAE / RMSE / MAPE(\%) & MAE / RMSE / MAPE(\%) \\ 
		\textbf{MegaCRN} & 1.93\textpm0.03 / 3.25\textpm0.07 / 4.63\textpm0.10 & 1.99\textpm0.03 / 3.53\textpm0.06 / 4.85\textpm0.09 & 2.60\textpm0.03 / 4.95\textpm0.08 / 6.70\textpm0.12 \\
		\midrule
		\multirow{2}{*}{\textbf{PEMS-BAY}} & 15min / horizon 3 & 30min / horizon 6 & 60min / horizon 12 \\
		 & MAE / RMSE / MAPE(\%) & MAE / RMSE / MAPE(\%) & MAE / RMSE / MAPE(\%) \\ 
		\textbf{MegaCRN} & 0.78\textpm0.02 / 1.40\textpm0.05 / 1.52\textpm0.05 & 0.85\textpm0.02 / 1.63\textpm0.04 / 1.68\textpm0.05 & 1.16\textpm0.03 / 2.56\textpm0.05 / 2.44\textpm0.06 \\
		\midrule
		\multirow{2}{*}{\textbf{Shuto-Expy}} & 10min / horizon 1 & 30min / horizon 3 & 60min / horizon 6 \\
		 & MAE / RMSE / MAPE(\%) & MAE / RMSE / MAPE(\%) & MAE / RMSE / MAPE(\%) \\ 
		\textbf{MegaCRN} & 3.47\textpm0.08 / 5.60\textpm0.13 / 15.77\textpm0.29 & 3.32\textpm0.13 / 5.55\textpm0.18 / 15.69\textpm0.62 & 4.53\textpm0.19 / 7.13\textpm0.23 / 20.51\textpm0.56 \\
		\bottomrule
	\end{tabular*}
\end{table*}
\begin{table*}[h]
    \footnotesize
	\centering
	\caption{Forecasting Performance in Incident Situations}
	\label{tab:incidentsituations}
	\begin{tabular*}{16.3cm}{@{\extracolsep{\fill}}cccccccccc}
		\hline
		Shuto-Expy & \multicolumn{3}{c}{10min / horizon 1} &
		\multicolumn{3}{c}{30min / horizon 3} &
		\multicolumn{3}{c}{60min / horizon 6} \\
		\cline{2-4} \cline{5-7} \cline{8-10}
		(Incidents) & 
		\multicolumn{1}{c}{MAE} & 
		\multicolumn{1}{c}{RMSE} &
		\multicolumn{1}{c}{MAPE} &
		\multicolumn{1}{c}{MAE} & 
		\multicolumn{1}{c}{RMSE} &
		\multicolumn{1}{c}{MAPE} &
		\multicolumn{1}{c}{MAE} & 
		\multicolumn{1}{c}{RMSE} &
		\multicolumn{1}{c}{MAPE} \\
		\hline
		HA\cite{li2018diffusion} & 14.72 & 19.38 & 87.23\% & 13.76 & 18.44 & 78.03\% & 12.85 & 17.47 & 68.87\% \\
		STGCN\cite{yu2018spatio} & 9.50 & 15.26 & 38.62\% & 11.51 & 17.74 & 51.77\% & 12.72 & 18.93 & 52.84\%\\
		DCRNN\cite{li2018diffusion} & 9.94 & 15.46 & 42.79\% & 12.11 & 18.10 & 52.90\% & 12.77 & 18.76 & 56.46\%\\
		GW-Net\cite{wu2019graph} & 9.52 & 15.12 & 40.08\% & 11.05 & 16.98 & 47.91\% & 10.99 & 17.00 & 46.56\% \\
		STTN\cite{xu2020spatial} & 9.31 & 14.59 & 40.58\% & 10.93 & 16.63 & 48.11\% & 11.34 & 17.22 & 51.08\%\\
		GMAN\cite{zheng2020gman} & 10.17 & 15.87 & 45.68\% & 11.78 & 17.91 & 51.53\% & 12.00 & 18.02 & 52.43\%\\
		MTGNN\cite{wu2020connecting} & 9.23 & 14.61 & 39.71\% & 10.84 & 16.97 & 47.14\% & 10.81 & 16.93 & 46.40\% \\
		StemGNN\cite{cao2020spectral} & 9.95 & 15.32 & 43.08\% & 12.25 & 17.97 & 54.61\% & 12.78 & 18.53 & 58.03\%\\
		AGCRN\cite{bai2020adaptive} & 9.38 & 14.51 & 41.62\% & 10.84 & 16.70 & 48.34\% & 10.95 & 16.98 & 48.25\% \\
		CCRNN\cite{ye2021coupled} & 9.40 & 14.78 & 39.66\% & 11.16 & 17.14 & 49.84\% & 11.27 & 17.40 & 51.05\%\\
		GTS\cite{shang2021discrete} & - & - & - & - & - & - & - & - & - \\
		PM-MemNet\cite{lee2021learning}  & 9.49 & 14.62 & 41.63\% & 10.92 & 16.90 &  46.98\% & 10.82 & 17.11 & 45.94\% \\
		\textbf{MegaCRN (Ours)} & \textbf{4.99} & \textbf{7.87} & \textbf{22.30\%} & \textbf{4.61} & \textbf{7.21} & \textbf{19.24\%} & \textbf{6.21} & \textbf{9.39} & \textbf{25.37\%} \\
		\hline
	\end{tabular*}
\end{table*}

\noindent\textbf{Preprocessing.}
The original data were aggregated over a time interval of 10 minutes by computing the average speed for each road link. Thus, there is a total of 144 speed-value points for every road link in a day. For the missing values in the intervals which were free of incidents, we replace the values using the mean speed observed during the same time interval on other days in that month for each individual road link. Here, we take care to use the same type of weekday to compute the mean speed, i.e., weekday for the weekday, and weekend for the weekend, to avoid any weekday bias. Moreover, only incident-free time intervals are considered for better accuracy. The remaining missing values were filled up using forward-fill method of Pandas. Lastly, a Boolean type value (i.e., incident occurred or not) was also created from original incident data for every 10-minute time interval corresponding to each road link.

\section{Quantitative Evaluation}
\subsection{Error Bars}
In Table \ref{tab:errorbars}, We report the error bars of our model (MegaCRN) on three datasets. Specifically, we run the model on each dataset 10 times and calculate their means and 95\% confidence intervals using  t-distribution. Note that in our main paper we listed the results of the baselines reported in their original papers and the best results of our model. Based on these error bars, even with upper confidence interval (CI), our model still can achieve big advantages over the baselines.

\subsection{Forecasting Performance on Incident Samples}

\begin{figure*}[h]
    \centering
    \begin{minipage}{0.495\linewidth}
        \centering
        \includegraphics[width=1.0\linewidth]{./figure/3cases_step0.png}
        \caption{Three incident cases (horizon 1).}
        \label{fig:3cases_step0}
    \end{minipage}
    \begin{minipage}{0.495\linewidth}
        \centering
        \includegraphics[width=1.0\linewidth]{./figure/3cases_step5.png}
        \caption{Three incident cases (horizon 6).}
        \label{fig:3cases_step5}
    \end{minipage}
\end{figure*}

Essentially, spatial-temporal data stream like urban traffic flow is a mixture of normality and abnormality. To verify the forecasting performance of the models during the abnormal situations, we further collect traffic incident information (e.g., congestion, regulation, accident, etc.) that happened in Shuto-Expy and match them to each link-timeslot pair in a consistent form $\mathbb{R}^{T\times N}$  with the traffic speed. According to the 0-1 incident flag (happen or not), we extract the testing samples that encountered an incident at the 1st prediction horizon (i.e., 10 minutes ahead). Table \ref{tab:incidentsituations} shows the forecasting accuracy averaged across all 6 horizons on these incident samples. Compared with Table 3 in the main paper, the errors become much higher than the normal-abnormal mixed situations as expected. But still, our model achieved the best records even in such incident situations, which demonstrated the superior adaptability over the baselines. In incident situations, our model improved the baselines by 49\%$\Delta$\textit{MAE} and 49\%$\Delta$\textit{RMSE} on average, which is even larger than the normal-abnormal mixed evaluations (40\%$\Delta$\textit{MAE} and 39\%$\Delta$\textit{RMSE}). This demonstrates the superior adaptability of our model over the SOTAs for non-stationary normal-abnormal mixed traffic forecasting. 

\begin{table*}[h]
    \small
    \centering
	\caption{Ablation Test at Representative Horizons}
	\label{tab:ablationstudy-horizon}
	\begin{tabular*}{16.3cm}{@{\extracolsep{\fill}}cccccccccc}
		\hline
		\multirow{2}{*}{METR-LA} & \multicolumn{3}{c}{15min / horizon 3} &
		\multicolumn{3}{c}{30min / horizon 6} &
		\multicolumn{3}{c}{60min / horizon 12} \\
		\cline{2-4} \cline{5-7} \cline{8-10}
		 & 
		\multicolumn{1}{c}{MAE} & 
		\multicolumn{1}{c}{RMSE} &
		\multicolumn{1}{c}{MAPE} &
		\multicolumn{1}{c}{MAE} & 
		\multicolumn{1}{c}{RMSE} &
		\multicolumn{1}{c}{MAPE} &
		\multicolumn{1}{c}{MAE} & 
		\multicolumn{1}{c}{RMSE} &
		\multicolumn{1}{c}{MAPE} \\
		\hline
		Adaptive & 2.84 & 5.44 & 7.40\% & 3.27 & 6.51 & 8.95\% & 3.80 & 7.74 & 10.86\% \\
		Memory & 2.82 & 5.41 & 7.38\% & 3.23 & 6.45 & 8.92\% & 3.74 & 7.65 & 10.78\% \\
		Momentary & 2.35 & 4.09 & 5.87\% & 2.43 & 4.31 & 6.19\% & 3.13 & 5.95 & 8.36\% \\
		\textbf{MegaCRN} & \textbf{1.90} & \textbf{3.18} & \textbf{4.57\%} & \textbf{1.96} & \textbf{3.46} & \textbf{4.79\%} & \textbf{2.60} & \textbf{4.86} & \textbf{6.70\%} \\
		\midrule
		\multirow{2}{*}{PEMS-BAY} & \multicolumn{3}{c}{15min / horizon 3} &
		\multicolumn{3}{c}{30min / horizon 6} &
		\multicolumn{3}{c}{60min / horizon 12} \\
		\cline{2-4} \cline{5-7} \cline{8-10}
		 & 
		\multicolumn{1}{c}{MAE} & 
		\multicolumn{1}{c}{RMSE} &
		\multicolumn{1}{c}{MAPE} &
		\multicolumn{1}{c}{MAE} & 
		\multicolumn{1}{c}{RMSE} &
		\multicolumn{1}{c}{MAPE} &
		\multicolumn{1}{c}{MAE} & 
		\multicolumn{1}{c}{RMSE} &
		\multicolumn{1}{c}{MAPE} \\
		\midrule
		Adaptive & 1.36 & 2.89 & 2.84\% & 1.71 & 3.85 & 3.83\% & 2.05 & 4.64 & 4.86\% \\
		Memory & 1.35 & 2.86 & 2.82\% & 1.69 & 3.82 & 3.82\% & 2.02 & 4.60 & 4.75\% \\
		Momentary & 0.99 & 1.87 & 1.99\% & 1.02 & 1.96 & 2.08\% & 1.36 & 2.93 & 2.98\% \\
		\textbf{MegaCRN} & \textbf{0.75} & \textbf{1.32} & \textbf{1.44\%} & \textbf{0.82} & \textbf{1.58} & \textbf{1.59\%} & \textbf{1.10} & \textbf{2.43} & \textbf{2.27\%} \\
		\midrule
		\multirow{2}{*}{Shuto-Expy} & \multicolumn{3}{c}{10min / horizon 1} &
		\multicolumn{3}{c}{30min / horizon 3} &
		\multicolumn{3}{c}{60min / horizon 6} \\
		\cline{2-4} \cline{5-7} \cline{8-10}
		 & 
		\multicolumn{1}{c}{MAE} & 
		\multicolumn{1}{c}{RMSE} &
		\multicolumn{1}{c}{MAPE} &
		\multicolumn{1}{c}{MAE} & 
		\multicolumn{1}{c}{RMSE} &
		\multicolumn{1}{c}{MAPE} &
		\multicolumn{1}{c}{MAE} & 
		\multicolumn{1}{c}{RMSE} &
		\multicolumn{1}{c}{MAPE} \\
		\hline
		Adaptive & 5.87 & 9.19 & 25.35\% & 6.53 & 10.41 & 29.66\% & 6.99 & 11.21 & 31.75\% \\
		Memory & 5.88 & 9.19 & 24.91\% & 6.57 & 10.49 & 30.22\% & 7.16 & 11.52 & 33.02\% \\
		Momentary & 5.11 & 7.81 & 20.82\% & 5.16 & 7.94 & 20.92\% & 5.94 & 9.07 & 25.34\% \\
		\textbf{MegaCRN} & \textbf{3.47} & \textbf{5.67} & \textbf{16.16\%} & \textbf{3.48} & \textbf{5.82} & \textbf{16.37\%} & \textbf{4.54} & \textbf{7.16} & \textbf{20.52\%} \\
		\hline
	\end{tabular*}
\end{table*}

We further verify the forecasting performance in incident situations (e.g., congestion, regulation, and accident) with three cases as listed in Figure \ref{fig:3cases_step0}$\sim$\ref{fig:3cases_step5}.
In each row, we plot the ground-truth traffic speed in one day (with accident happen) and the prediction results of two baselines (CCRNN~\cite{ye2021coupled} and GW-Net~\cite{wu2019graph}) and our model (MegaCRN) for one specific road link. Red blocks show the time intervals when accidents happened. Figure \ref{fig:3cases_step0} shows the results of the 1st horizon (10 minutes ahead prediction), while Figure \ref{fig:3cases_step5} shows the corresponding results (the same road links with Figure \ref{fig:3cases_step0}) of the 6th horizon. Through Figure \ref{fig:3cases_step0}$\sim$\ref{fig:3cases_step5}, we could find that MegaCRN achieves much better results in both normal and abnormal situations (w/ and w/o accidents). When the accident happened and the speed suddenly dropped down, MegaCRN could quickly adapt to the sudden speed change, while the two baselines have some delay or even fail to do so. Especially, compared with 10 minutes ahead forecasting in Figure \ref{fig:3cases_step0}, GW-Net and CCRNN performed much worse for 60 minutes ahead forecasting. GW-Net could not even catch the sudden speed change for such long-horizon forecasting in accident situations as shown in Figure \ref{fig:3cases_step5}. By contrast, MegaCRN could achieve almost the same performance for horizon 1 and 6, which further verified the superior adaptability of our model. 

\subsection{Ablation Test}
Due to the space limitation, we reported the results of ablation test across all horizons as Table 4 in the main paper. Here, we report the results for the representative horizons as Table \ref{tab:ablationstudy-horizon}. The conclusions are consist between the two tables, so we omit the elaboration here.
\subsection{Efficiency Evaluation}
\begin{figure}[h]
	\centering
	\includegraphics[width=1.0\linewidth]{./figure/scatters.png}
	\caption{Efficiency Evaluation}
	\label{fig:efficiency}
\end{figure}
We also evaluate the efficiency of our model by comparing with the-state-of-the-arts. Here we just report the results on Shuto-Expy, because the spatial domain of our data is 5$\sim$9 times larger than the benchmarks. Two scatter plots are shown as Figure \ref{fig:efficiency}. In the left figure, the x-axis is the total number of model parameters and the y-axis is the total training time in minutes (epochs needed to converge $\times$ running time per epoch). In the right figure, the x-axis of is the total number of parameters and the y-axis is the overall MAE. We can see that our model has the second-fewest parameters (i.e., merely 64,217) and the fourth least total training time, but the smallest overall MAE. For a large-scale dataset like Shuto-Expy, our model could be very memory-efficient. In contrast, some models, especially Transformer-based models including GMAN~\cite{zheng2020gman} and STTN~\cite{xu2020spatial}, are very memory/time-consuming due to the dot-product operation on big tensor. Although our model tends to need more epochs to converge, each round of training could be finished in very little time. To sum up, our model can achieve the state-of-the-art precision while keeping comparatively efficient.

\subsection{Hyperparameter Study} \label{sec:experiment-hyperparameter}
\begin{figure}[h]
	\centering
	\includegraphics[width=1.0\linewidth]{./figure/Hyperparameter-3.png} 
	\caption{Hyperparameter Study}
	\label{fig:Hyperparameter}
\end{figure}
We also conduct several experiments about the hyperparameters and show how they effect the model performance. We select the most important two hyperparameters, namely the number of meta-nodes and the dimension of each meta-node. We test the meta-node number out of \{2, 4, 6, 8, 10\} and the dimension out of \{8, 16, 32, 64\} on all three datasets and plot the MAEs across all horizons in Figure \ref{fig:Hyperparameter}. Through it, we could find that keeping increasing the number and dimension could not always assure the performance gain on all three datasets. Because the number of meta-nodes (i.e., hidden prototypes) implied in each dataset depends on the data complexity. Generally, 10 meta-nodes with 32 dimensions can work well enough on all three datasets while keeping the model tiny and efficient. Specifically, by taking Shuto-Expy dataset as an example, 10 meta-nodes/32 dimensions bring us a total of 64,217 parameters while 10 meta-nodes/64 dimensions bring us 112,601 parameters. Correspondingly, the total training time also increased by around 45\% (from 4,200 seconds to 6,100 seconds) by enlarging the dimension from 32 to 64. 

\section{Qualitative Evaluation}
\subsection{Spatio-Temporal Disentanglement}
We further demonstrate the disentangling effects of Meta-Graph approach in three domains, including space, time, and memory (Meta-Node Bank).

\begin{figure}[h]
    \centering
    \begin{minipage}{0.495\linewidth}
        \centering
        \includegraphics[width=1.0\linewidth]{./figure/Spaio-Temporal Disentangling-Apndx-1.png}
        \caption{Distributions of 138 Toll Gates and 1220 Road Links in Cluster 1}
        \label{fig:st-dis-1}
    \end{minipage}
    \begin{minipage}{0.495\linewidth}
        \centering
        \includegraphics[width=1.0\linewidth]{./figure/Spaio-Temporal Disentangling-Apndx-2.png}
        \caption{Distributions of 138 Toll Gates and 623 Road Links in Cluster 2}
        \label{fig:st-dis-2}
    \end{minipage}
\end{figure}

\noindent\textbf{Spatial Disentanglement.} Figure \ref{fig:st-dis-1} and \ref{fig:st-dis-2} illustrate the locations of road links in discovered clusters (refer to \textbf{Section 5.3}) and toll gates. Compared with road links in cluster 1 (in blue), the ones in cluster 2 (in red) show a stronger clustering with toll gates, which is quantitatively validated by Ripley's K \cite{ripley1976second}, a measure of stationary point patterns:
\begin{equation} \label{eq:cross-k}
	K(r) = \frac{E[r]}{\lambda}
\end{equation}
where a $K$ value is a product of the inverse of one type of points' intensity ($\lambda$) and the expectation of additional points within a fixed distance ($E[r]$). Spatial clustering or dispersion between two types of points can be evaluated by graphically comparing the K function of a desired type to corresponding K value of complete spatial randomness patterns (in black dotted curves). A constantly higher position of K function of cluster 2 indicates its spatial clustering (closely distributed) with toll gates. This observation demonstrates that the learnt clusters by Meta-Graph can be spatially interpreted, indicating the spatial disentangling power of our approach.
\begin{figure}[h]
    \centering
    \begin{minipage}{1.0\linewidth}
    	\centering
    	\includegraphics[width=1.0\linewidth]{./figure/Spaio-Temporal Disentangling-Apndx2-1.png}
    	\caption{Time Series Histogram of Daily Average Traffic Speed in \textit{Cluster 1}}
    	\label{fig:TSH1}
	\end{minipage}
    \begin{minipage}{1.0\linewidth}
    	\centering
    	\includegraphics[width=1.0\linewidth]{./figure/Spaio-Temporal Disentangling-Apndx2-2.png}
    	\caption{Time Series Histogram of Daily Average Traffic Speed in \textit{Cluster 2}}
    	\label{fig:TSH2}
    \end{minipage}
\end{figure}

\noindent\textbf{Temporal Disentanglement.} In Figure \ref{fig:TSH1} and \ref{fig:TSH2}, we further plot the time series histograms of daily average traffic speed in cluster 1 and cluster 2 respectively. Clearly, the former shows a more typical curve of traffic speed (roads impacted by rush-hours in varied ways), while the latter features a lower average speed but a higher variation, plus no daytime pattern. This observation is actually consistent with the previous one in spatial disentanglement, interestingly. A lot of accelerating and decelerating happen near toll gates and interchanges, which explains what we observes in cluster 2 (Figure \ref{fig:TSH2}). These distinctive patterns demonstrate the temporal disentangling power of Meta-Graph.
\begin{figure}[h]
	\centering
	\begin{minipage}{1.0\linewidth}
		\centering
    	\includegraphics[width=1.0\linewidth]{./figure/Spaio-Temporal Disentangling-Apndx3-1.png}
    	\caption{\textit{Cluster 1}'s Average Query Scores ($\bar{a}_j$) in Meta-Node Bank Towards Weekend}
    	\label{fig:MEM1}
    \end{minipage}
	\begin{minipage}{1.0\linewidth}
    	\centering
    	\includegraphics[width=1.0\linewidth]{./figure/Spaio-Temporal Disentangling-Apndx3-2.png}
    	\caption{\textit{Cluster 2}'s Average Query Scores ($\bar{a}_j$) in Meta-Node Bank Towards Weekend}
    	\label{fig:MEM2}
	\end{minipage}
\end{figure}

\textbf{Disentanglement in Memory Space.} We visualize Meta-Node Bank's average query scores in the learnt clusters (1 in Figure \ref{fig:MEM1}, 2 in Figure \ref{fig:MEM2}). We choose Fridays, Saturdays and Sundays' average because they show most characteristic patterns (other weekdays are Friday-alike). Two major observations can be made: (1) Different memory items characterize different clusters. While item 0 and 9 are mainly responsible for cluster 1, item 2 and 7 are active for cluster 2. This means the disentanglement roots in Meta-Node Bank, which memorizes different patterns in different items. (2) Weekday and weekend are distinguished by querying memory differently. In cluster 1, higher weight on item 0 indicates a decelerating, while higher weight on item 9 indicates an accelerating. Thus, morning rush-hour manifests on Fridays, dims on Saturdays, and vanishes on Sundays; evening rush-hour lasts after 18:00 on Fridays, ends before 18:00 on Saturdays, and fades on Sundays. In cluster 2, querying patterns seem rather constant and do not show weekday-weekend pattern, which is consistent with Figure \ref{fig:TSH2}.
\begin{figure}[h]
    \centering
    \begin{minipage}{1.0\linewidth}
	    \centering
    	\includegraphics[width=1.0\linewidth]{./figure/Spaio-Temporal Disentangling-Apndx3-3.png}
    	\caption{Average Query Scores ($\bar{a}_j$) in Meta-Node Bank ($\kappa_1$=0.01, $\kappa_2$=0.01)}
    	\label{fig:MEM3}
    \end{minipage}
    \begin{minipage}{1.0\linewidth}
    	\centering
    	\includegraphics[width=1.0\linewidth]{./figure/Spaio-Temporal Disentangling-Apndx3-4.png}
    	\caption{Average Query Scores ($\bar{a}_j$) in Meta-Node Bank ($\kappa_1$=0, $\kappa_2$=0)}
    	\label{fig:MEM4}
	\end{minipage}
\end{figure}

Moreover, we evaluate the two constraints (consistency loss $\mathcal{L}_{con1}$ and contrastive loss $\mathcal{L}_{con2}$ in \textbf{Section 4.3}) used for regulating memory parameters by comparing two cases, training with them ($\kappa_1$=0.01, $\kappa_2$=0.01 in Figure \ref{fig:MEM3}) and without them ($\kappa_1$=0, $\kappa_2$=0 in Figure \ref{fig:MEM4}). Another two observations can be made: (1) Figure \ref{fig:MEM3} can be seen as a combination of Figure \ref{fig:MEM1} and Figure \ref{fig:MEM2}, which means each memory item is relatively independent in storing spatio-temporal prototypes (when we train MegaCRN with the constraints). (2) Comparing Figure \ref{fig:MEM4} with \ref{fig:MEM3}, we can tell the information is condensed, majorly into memory item 3 (constantly higher querying weight) when the constraints are not applied. This observation confirms the distinguishing power brought by the parameter constraints for implementation of sparse attention mechanism.

\subsection{Incident-Awareness}
\begin{figure*}[h]
    \centering
    \begin{minipage}{1.0\linewidth}
    	\centering
    	\includegraphics[width=1.0\linewidth]{./figure/Incident-Aware-new-2.png}
    	\caption{Additional Case Study 1 on Incident Awareness}
    	\label{fig:case-incident-2}
    \end{minipage}
    \begin{minipage}{1.0\linewidth}
    	\centering
    	\includegraphics[width=1.0\linewidth]{./figure/Incident-Aware-new-3.png}
    	\caption{Additional Case Study 2 on Incident Awareness}
    	\label{fig:case-incident-3}
    \end{minipage}
    \begin{minipage}{1.0\linewidth}
        \centering
    	\includegraphics[width=1.0\linewidth]{./figure/Incident-Aware-new-4.png}
    	\caption{Additional Case Study 3 on Incident Awareness}
    	\label{fig:case-incident-4}
	\end{minipage}
\end{figure*}

\begin{table*}[h]
    \small
	\centering
	\caption{Performance of Multi-Step Forecasting on Electricity Benchmark}
	\label{tab:electricity}
	\begin{tabular*}{16.0cm}{@{\extracolsep{\fill}}ccccccc}
		\hline
		\multicolumn{1}{c}{} & 
		\multicolumn{2}{c}{3 hours / horizon 3} &
		\multicolumn{2}{c}{6 hours / horizon 6} &
		\multicolumn{2}{c}{12 hours / horizon 12}
		\\
		\hline
        \multicolumn{1}{c}{\textbf{Model}} & 
		\multicolumn{1}{c}{RSE} & 
		\multicolumn{1}{c}{CORR} &
		\multicolumn{1}{c}{RSE} & 
		\multicolumn{1}{c}{CORR} & 
		\multicolumn{1}{c}{RSE} & 
		\multicolumn{1}{c}{CORR}
		\\
		\hline
		HA & 0.1307 & 0.8876 & 0.1307 & 0.8875 & 0.1307 & 0.8873 \\
		LSTNet\cite{lai2018modeling} & 0.1290 & 0.9203 & 0.1615	& 0.8872 & 0.1215 & 0.8999\\
		ST-Norm\cite{deng2021st} & 0.1222 & 0.9147 & 0.1290 & 0.8993 & 0.1333 & 0.8887\\
		GW-Net\cite{wu2019graph} & 0.1159 & 0.9256 & 0.1335 & 0.9130 & 0.1254 &	0.9064\\
		STTN\cite{xu2020spatial} & 0.1122 & 0.9233 & 0.1341	& 0.9076 & 0.1217 & 0.9038\\
		MTGNN\cite{wu2020connecting} & 0.1186 & 0.9290 & 0.1293 & 0.9180 & 0.1189 & 0.9132\\
		StemGNN\cite{cao2020spectral} & 0.1306 & 0.9100 & 0.1415 & 0.8925 & 0.1205 & 0.8939\\
		AGCRN\cite{bai2020adaptive} & 0.1047 & 0.9287 & 0.1158 & 0.9168 & 0.1156 & 0.9146 \\
		CCRNN\cite{ye2021coupled} & 0.1059 & 0.9165 & 0.1141 & 0.9044 & 0.1166 & 0.9018\\
		\textbf{MegaCRN (Ours)} & \textbf{0.0938} & \textbf{0.9421} & \textbf{0.1054} & \textbf{0.9338} & 0.1448 & \textbf{0.9249}\\
        \hline
	\end{tabular*}
\end{table*}

We further provide three more case studies on non-stationary traffic conditions in Figure \ref{fig:case-incident-2}, \ref{fig:case-incident-3}, and \ref{fig:case-incident-4}. Case 1 and 2 illustrate two scenarios where an incident on one road link impacts its adjacent roads. The dominant positions of upstream node 2 in Figure \ref{fig:case-incident-2} and node 6 in Figure \ref{fig:case-incident-3} are taken over right after traffic incidents. Case 3 shows an interesting instance where a disturbance occurs on a road link in cluster 2 (based on the memory query pattern). Here node 8 is an exit connected to node 7. At 21:10 on December 2nd, traffic speed on node 8 suddenly dropped to zero and recovered shortly. Although it was labeled as incident, we can tell it could actually be a sensor failure according to the shape of time series. In this case, MegaCRN has a smallest and prompt fluctuation compared with two baselines (GW-Net~\cite{wu2019graph} and CCRNN~\cite{ye2021coupled}). This case further proves the robustness and adaptivity of our approach under non-stationarity.

\section{Multivariate Time Series Forecasting}
Here, we test our model for Multivariate Time Series (MTS) Forecasting task based on the Electricity benchmark published by \cite{lai2018modeling}. The electricity consumption in kWh was recorded every 1 hour from 2012/1/1 to 2014/12/31 (total 26,304 timesteps) for N = 321 clients. The observation step is 12 (past 12 hours) and the prediction horizon is 12 (12 hours ahead), which is a long-term forecasting compared to our traffic forecasting. The settings are almost the same as our main experiment on the two traffic benchmarks. For example, the ratio for training, validation, and testing is 7:1:2. Z-score normalization was applied for each variable. Adam was used as the optimizer, where the learning rate was set to 0.001 and the batch size was set to 64. The optimizer would either be early-stopped if the validation error was converged within 10 epochs or be stopped after 200 epochs. L1 Loss is used as the loss function. Following the tradition of the previous works~\cite{lai2018modeling,wu2020connecting}, we use Root Relative Squared Error (RSE) and Empirical Correlation Coefficient (CORR) as the metrics. The range of value (0$\sim$764000) in Electricity dataset is very large, which means the fluctuation is much severer than the Traffic dataset. Thus, we make our model deeper by using two layers of RNNs in encoder and decoder respectively, each of which has 32 hidden states. Same with the main experiment, we also use 10 meta-nodes, each of which is a 32-dimension learnable vector. The two balancing factors $\kappa_1$ and $\kappa_2$ are both set to 0.01.

We list the experimental results in Table~\ref{tab:electricity}. It is worthy noting that some baselines used in the Traffic dataset including STGCN~\cite{yu2018spatio}, DCRNN~\cite{li2018diffusion}, GMAN~\cite{zheng2020gman}, and PM-MemNet\cite{lee2021learning} require the pre-defined graph (i.e., adjacency matrix), which is not available in this dataset, so we remove them from the baselines. Instead, we add two MTS SOTAs, namely LSTNet~\cite{lai2018modeling} and STNorm~\cite{deng2021st}. Through Table~\ref{tab:electricity}, we can see that MegaCRN could achieve the best performance for the relatively short horizons (3 and 6 hours ahead forecasting), but fail to outperform the baselines (still the best CORR but the worst RSE) for the long horizon (12 hours ahead forecasting). This demonstrates the limitation of our model, that is, unsatisfactory performance for long-term forecasting. 

\textbf{Limitation.} The reasons behind this are as follows: (1) Compared to the SOTAs such as ST-Norm~\cite{deng2021st}, GW-Net~\cite{wu2019graph}, and MTGNN~\cite{wu2020connecting} that are based on WaveNet~\cite{oord2016wavenet}, MegaCRN is built based upon an RNN-based encoder-decoder architecture that performs a standard step-wise decoding in the decoder. The nature of RNN brings us the limitation for long-term forecasting. That is also why WaveNet~\cite{oord2016wavenet} and Transformer~\cite{vaswani2017attention} are more widely used backbones for long-term forecasting. (2) In our current model, as illustrated in Figure 2 in our main paper, the learned meta-graph $\mathcal{G}'$ is shared across all of the horizons in the decoder. This is specially designed based on the assumption: the interactions among the road links will be relatively stationary within one hour, since the forecasting target for all the Traffic datasets is up to one hour. By letting the graph be shared across horizons, it helps reduce the number of parameters of our model and make it easier to be trained, as learning one single meta-graph is much more easier than learning multiple ones. The time interval of the Electricity data is 1 hour and we aim to conduct a multi-step forecasting from horizon 1 to 12. Apparently, the latent correlations among the variates vary a lot during the 12 hours, for which the shared meta-graph mechanism cannot work.

\section{Future Work} 
Correspondingly, we consider to improve our model from the following aspects: (1) Instead of RNN, we aim to implement the Spatio-Temporal Meta-Graph Learning based on  WaveNet~\cite{oord2016wavenet} or Transformer~\cite{vaswani2017attention} backbones. (2) We will explore ways of learning multiple meta-graphs for each step instead of the shared one. Moreover, in our current model, we only use one FC layer as the Hyper-Network to generate the meta-graph, where more advanced neural networks could be employed for further improvement.

\bibliography{aaai23}